\documentclass[10pt,twocolumn,letterpaper]{article}

\usepackage[pagenumbers]{iccv} 

\definecolor{iccvblue}{rgb}{0.21,0.49,0.74}
\usepackage[pagebackref,breaklinks,colorlinks,allcolors=iccvblue]{hyperref}

\usepackage{graphicx}
\usepackage{amsmath}
\usepackage{amssymb}
\usepackage{booktabs}
\usepackage{times}
\usepackage{epsfig}
\usepackage{multirow}
\usepackage{color}
\usepackage{arydshln}
\usepackage{float}
\usepackage{caption}
\usepackage{pifont}
\usepackage[pagebackref,breaklinks,colorlinks]{hyperref}

\usepackage[capitalize]{cleveref}
\crefname{section}{Sec.}{Secs.}
\Crefname{section}{Section}{Sections}
\Crefname{table}{Table}{Tables}
\crefname{table}{Tab.}{Tabs.}

\definecolor{best}{HTML}{bcf9bb}

\definecolor{secondbest}{HTML}{ffacac}

\def\paperID{4}
\def\confName{ICCVW}
\def\confYear{2025}

\title{Learning Camera-Agnostic White-Balance Preferences}

\author{\\[-10pt] 
Luxi Zhao \hspace{15pt} Mahmoud Afifi \hspace{15pt} Michael S. Brown \\[3pt]  
AI Center-Toronto, Samsung Electronics \\[3pt]  
{\tt\small \{lucy.zhao, m.afifi1, michael.b1\}@samsung.com}\\[-10pt]   
}

\begin{document}

\twocolumn[{%
\renewcommand\twocolumn[1][]{#1}%
\maketitle
\begin{center}
\vspace{-2mm}
\includegraphics[width=\textwidth]{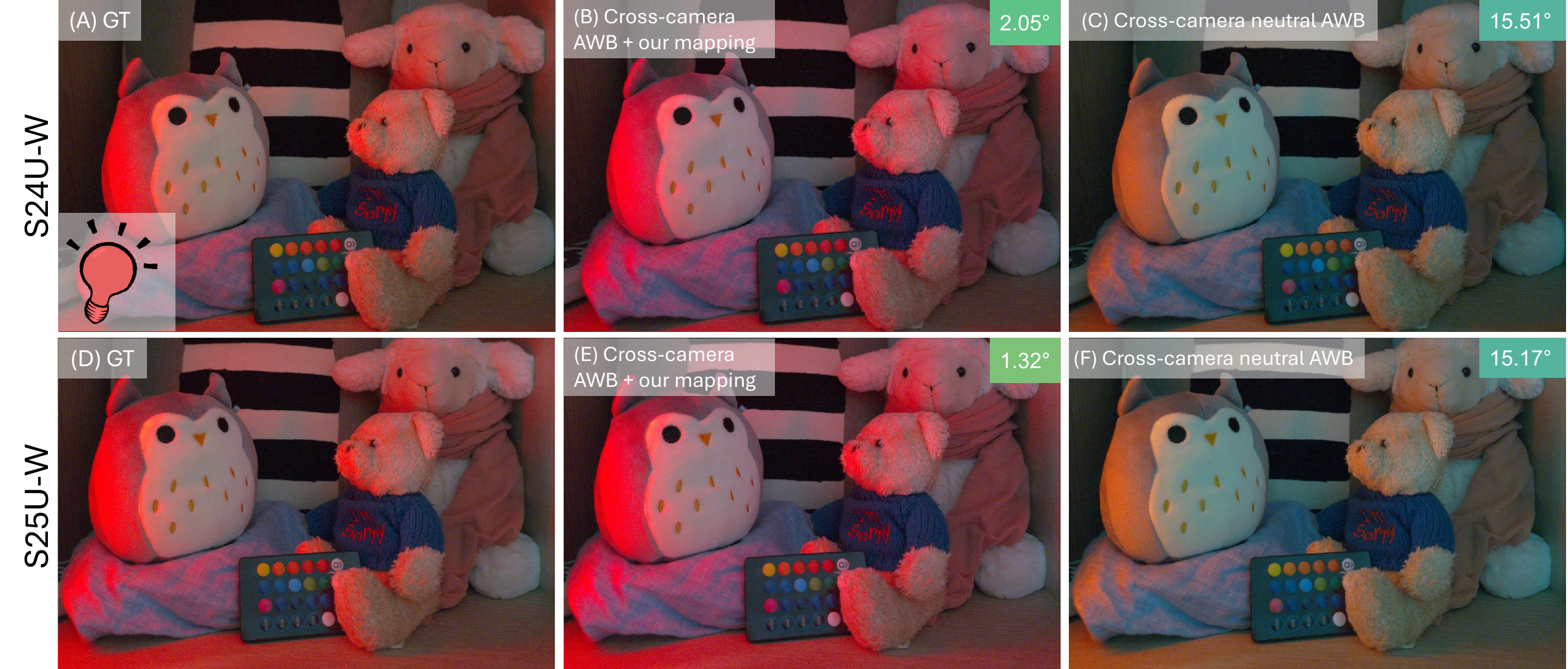}
\captionof{figure}{
Cross-camera AWB methods typically aim for neutral color correction (subfigures C and F). In practice, however, camera manufacturers often design AWB algorithms to reflect white-balance preferences that achieve a desired aesthetic (subfigures A and D). We introduce a camera-agnostic learnable mapping function that transforms the neutral illuminant predicted by cross-camera AWB algorithms into a target white-balance preference, enabling consistent rendering across different cameras (B and E). Shown are: (A) a scene captured by the Galaxy S24 Ultra wide (S24U-W) camera, rendered using the ``ground-truth'' white-balance preference; (B) the same scene corrected using our mapping trained on S24U-W data; (C) the same image corrected using C5 \cite{C5}'s neutral white balance; (D) the same scene captured by the Galaxy S25 Ultra wide (S25U-W) camera, rendered using the ``ground-truth'' preference; (E) the same S25U-W image corrected using our mapping trained on S24U-W; and (F) the same S25U-W image corrected using C5’s neutral white balance.\label{fig:teaser}}
\end{center}%
}]

\maketitle

\begin{abstract}
The image signal processor (ISP) pipeline in modern cameras consists of several modules that transform raw sensor data into visually pleasing images in a display color space. Among these, the auto white balance (AWB) module is essential for compensating for scene illumination. However, commercial AWB systems often strive to compute aesthetic white-balance preferences rather than accurate neutral color correction. While learning-based methods have improved AWB accuracy, they typically struggle to generalize across different camera sensors---an issue for smartphones with multiple cameras. Recent work has explored cross-camera AWB, but most methods remain focused on achieving neutral white balance. In contrast, this paper is the first to address aesthetic consistency by learning a post-illuminant-estimation mapping that transforms neutral illuminant corrections into aesthetically preferred corrections in a camera-agnostic space. Once trained, our mapping can be applied after any neutral AWB module to enable consistent and stylized color rendering across unseen cameras. Our proposed model is lightweight---containing only $\sim$500 parameters---and runs in just 0.024 milliseconds on a typical flagship mobile CPU. Evaluated on a dataset of 771 smartphone images from three different cameras, our method achieves state-of-the-art performance while remaining fully compatible with existing cross-camera AWB techniques, introducing minimal computational and memory overhead.
\end{abstract}

\section{Introduction}
\label{sec:intro}

Auto white balance (AWB) is a fundamental module in the camera image signal processor (ISP) pipeline \cite{delbracio2021mobile, brown2023color}, which converts raw sensor data into aesthetically pleasing images in a display color space. Ideally, the AWB module approximates color constancy, aiming to render object colors consistently regardless of the scene’s illumination \cite{agarwal2006overview}. Specifically, AWB attempts to map the scene’s illuminant color to the achromatic line in the camera’s raw space---that is, to align raw RGB values corresponding to illumination with the $R$ = $G$ = $B$ ``white'' line \cite{afifi2019color}.

However, in practice, commercial camera ISPs often incorporate aesthetic considerations that go beyond purely neutral white balancing. These preferences are influenced by photographic trends and do not always adhere to the white-light assumption \cite{afifi2019tcolor, afifi2020deep, ershov2023physically, scuello2004museum, hu2018exposure, cheng2016two}. In fact, neutral white balancing---which attempts to fully neutralize the lighting color in the scene---can sometimes produce images that appear unnatural. This is partly due to the human visual system’s incomplete chromatic adaptation \cite{incomplete_cat1, tominaga2014prediction}, which limits our ability to fully discount strong environmental lighting. Furthermore, camera manufacturers frequently bias the white balance away from neutral to reflect aesthetic preferences or maintain a brand-specific tonal style. These aesthetic adjustments cannot be captured by a neutral illuminant alone (see Fig.~\ref{fig:teaser}).

AWB typically involves two main steps: (1) illuminant estimation and (2) color correction \cite{gijsenij2011computational}. In the first step, the goal is to estimate a single color vector representing the color bias introduced by the scene’s illumination in the camera-captured raw image. This estimated vector---commonly referred to as the illuminant color---is then used in the second step, which applies global scaling to the $R$, $G$, and $B$ channels of the raw image to compensate for the color bias \cite{afifi2019color}. Most prior work focuses on the illuminant estimation step and frames the problem within the context of color constancy, aiming to neutralize the effect of the scene’s illumination \cite{FFCC, CCC, wGE, GW, BoCF, SoG, GI, BMVC1, MSGP, NUS}. These methods seek to accurately predict the global illuminant color in the camera's raw color space to achieve a neutral white balance.

\begin{figure}[t]
\centering
\includegraphics[width=\linewidth]{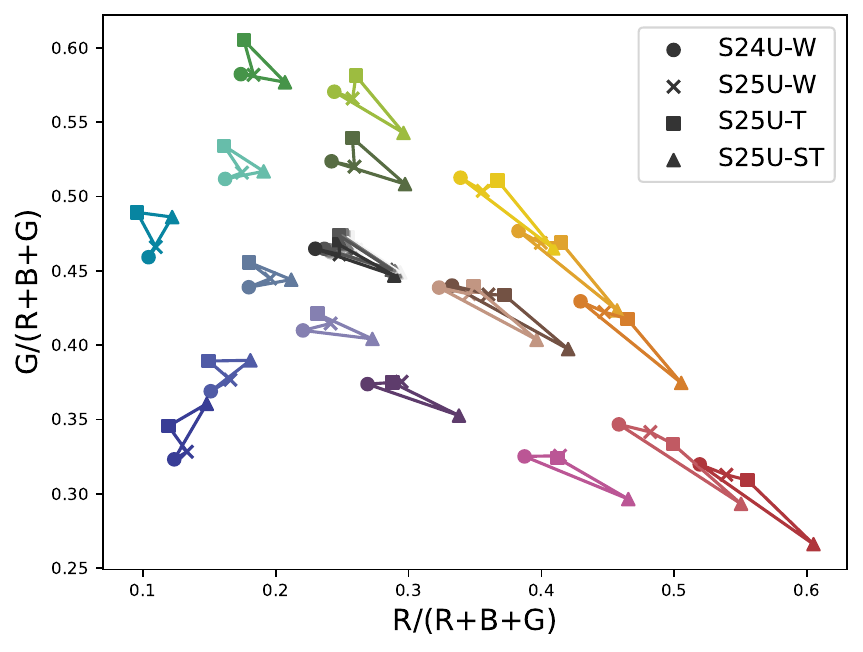}
\caption{Each point in the figure represents a raw color from a Macbeth color checker. For the same object under identical illumination, different sensors produce different raw values due to variations in spectral response and other camera characteristics. Without special handling, models must be trained per sensor to accurately predict raw illuminants. Shown are examples from the Samsung S24 Ultra’s wide camera (S24U-W), Samsung S25 Ultra’s wide camera (S25U-W), Samsung S25 Ultra’s telephoto camera (S25U-T), and Samsung S25 Ultra’s super-telephoto camera (S25U-ST).
\label{fig:raw_colorchart}}
\end{figure}

Recent state-of-the-art illuminant estimation methods are learning-based and require training on paired datasets consisting of raw images captured by a single camera along with ground-truth illuminants \cite{FC4, FFCC, lo2021clcc, hernandez2020multi, afifi2025optimizing}. These ground truths are typically obtained using a calibration object (e.g., a color chart) placed in the scene to accurately capture the scene illumination. Such models are inherently camera-specific and struggle to generalize to new cameras with different sensor spectral sensitivities. This is because each camera interprets raw RGB values differently (see Fig.~\ref{fig:raw_colorchart}), making a model trained on one sensor unreliable when applied to another \cite{SIIE, kim2025ccmnet}.

To deploy AWB models across devices, manufacturers would need to retrain or fine-tune the model for each new sensor, which is impractical---especially for mobile manufacturers who release multiple devices per year, each with potentially different cameras. To address this, recent work has attempted to retain the power of learning-based methods while improving their generalization without requiring retraining \cite{SIIE, C4, C5, kim2025ccmnet}. These models, termed {\it cross-camera AWB}, are typically trained on data from multiple cameras and incorporate architectural or training strategies that help the model adapt to new sensors. However, to ensure consistent supervision across sensors---each with its own raw RGB color space---these models are often restricted to using neutral illuminants as training targets. While practical, this constraint limits their ability to model aesthetic variations tailored to visual preferences or brand-specific styles, highlighting the need for a cross-camera approach that goes beyond color constancy to account for both perceptual and stylistic consistency.

This motivated us to develop a method that enables accurate, preference-aware white balance across cameras. Specifically, we propose a technique that augments existing cross-camera AWB methods by introducing an additional learnable mapping function to model target white-balance preferences. Our method is designed to be camera-agnostic, meaning that once the mapping is trained, it can be applied to unseen cameras without re-training, making it highly practical for camera manufacturers.

To evaluate our method’s ability to produce consistent illuminant predictions with brand-specific or aesthetically-preferred biases across different sensors, we collected a test set from devices made by the same manufacturer, assuming that their onboard AWB systems reflect a consistent aesthetic preference. We used these in-camera AWB illuminant outputs as ground truth to train and evaluate our method. Additionally, we tested our model on the S24 dataset \cite{afifi2025time}, which includes illuminants manually annotated by a professional photographer to reflect aesthetic preferences. Our results show consistent improvements---both qualitatively and quantitatively---across testing sets captured with different cameras, demonstrating the method's potential for real-world deployment.

\paragraph{Contribution}
In this paper, we propose a method for incorporating white-balance preferences into cross-camera AWB frameworks. Our approach introduces a post-illuminant-estimation mapping function trained to convert neutral illuminants predicted by existing cross-camera AWB models into manufacturer-preferred illuminants within a camera-agnostic space. This allows the method to generalize effectively to unseen cameras with varying characteristics. We evaluate our approach on a dataset of 771 smartphone images captured by three cameras and show consistent improvements over standard cross-camera AWB methods in achieving the desired white balance style. Our model is lightweight, containing only $\sim$500 tunable parameters, and runs in under 0.05 milliseconds on a typical flagship mobile CPU.

\section{Related Work}
\label{sec:related-work}
As stated earlier, most prior research has focused on accurately predicting the scene’s illuminant color from a given image or auxiliary inputs for neutral AWB \cite{GI, FFCC, FC4, C5, two-camera, afifi2025optimizing}. Going beyond this, our method targets cross-camera illuminant estimation while explicitly modeling target white-balance preferences through a post-illuminant estimation mapping. To provide context, we briefly review related work in three categories: (1) camera-specific illuminant estimators, (2) cross-camera illuminant estimators, and (3) color mapping techniques related to white balance and in-camera color correction.

\subsection{Camera-Specific Illuminant Estimators}
Camera-specific illuminant estimators are learning-based models trained on paired datasets captured by a single camera (e.g., \cite{CCC, FFCC, APAP, FC4, BoCF, PCC, BMVC1, DSNET, two-camera, afifi2025optimizing, afifi2025time}). These models learn a mapping from input images (or derived color features), optionally combined with additional cues (e.g., \cite{two-camera, afifi2025optimizing, afifi2025time}), to ground-truth illuminant colors obtained from a calibration object placed in the scene. These ground-truth illuminants are used to supervise models for neutral AWB, which aims to normalize both sensor sensitivity bias and the color cast introduced by scene illumination. As a result, most of these methods are inherently designed for neutral white balancing, due to the nature of how their ground-truth data is collected. A recent exception is the S24 dataset \cite{afifi2025time}, which introduces ground-truth labels based on aesthetic preference, manually curated by a professional photographer. 

Regardless of the supervision type, these models typically struggle to generalize to new cameras with different sensor characteristics \cite{SIIE}. This is because each sensor has its own raw RGB color space, and the same RGB value can correspond to different physical colors across sensors \cite{afifi2021semi}. Consequently, re-training or fine-tuning is often required to adapt such models to new sensors---an impractical requirement in real-world scenarios, particularly in the smartphone industry, where manufacturers release multiple devices with varying camera modules each year.

\subsection{Cross-Camera Illuminant Estimators}
Cross-camera illuminant estimators are designed to deliver consistent performance regardless of the testing camera’s characteristics. These methods fall into two categories: (1) inherently camera-agnostic approaches, such as statistical-based methods (e.g., \cite{GW, NUS, GI, wGE, maxRGB, MSGP}), which estimate the illuminant directly from the image content without relying on learned priors; and (2) learning-based models that incorporate design elements to enhance generalization to unseen cameras at test time (e.g., \cite{SIIE, C4, C5, kim2025ccmnet}).

Recent state-of-the-art cross-camera learning-based methods often rely on additional inputs beyond the image to help the model adapt to new sensors. For instance, C5 \cite{C5} requires auxiliary images captured by the test-time camera to better adapt to the color space of the testing camera. CCMNet \cite{kim2025ccmnet} leverages the camera’s color correction matrices (CCMs)---also known as color space transformation (CST) matrices---typically calibrated within the ISP, to aid adaptation to the test camera’s color space.

While these methods achieve promising results across cameras, they are generally trained on data from multiple sensors with diverse characteristics and are limited to neutral AWB. This restriction arises because training on preferred white balance would require ground-truth preference data to be consistent across all cameras---a condition that is difficult to guarantee. Collecting such data would involve a highly controlled and labor-intensive process to ensure uniform aesthetic preferences across different sensors, making it impractical for the large-scale datasets required to train these models.

\subsection{Color Mapping}
Camera ISPs apply various color mapping transformations to render raw sensor data into display-ready images \cite{delbracio2021mobile}. One such transformation is performed immediately after white balancing, using pre-calibrated CST matrices \cite{brown2023color}. These matrices are typically calibrated in the lab during manufacturing using color charts, and are designed to map white-balanced raw data into a canonical, device-independent color space (e.g., CIE XYZ) \cite{finlayson2015color, afifi2021cie}. 

Color mapping has also been utilized within illuminant estimation and white balance methods. For example, sensor sharpening \cite{finlayson1994spectral} applies a transformation to raw images to improve illuminant estimation accuracy. SIIE \cite{SIIE} learns a 3×3 invertible matrix to map raw images from various camera color spaces into a unified ``working'' space to enhance cross-camera generalization. APAP \cite{APAP} introduces a post-illumination-estimation camera-specific mapping to refine initial illuminant estimates. The work in \cite{afifi2020interactive} proposes a rectification mapping function that projects initial illuminant estimates into a higher-order polynomial space to address non-linear color shifts caused by in-camera white balancing errors.  While our method also learns a post-illuminant-estimation mapping, to the best of our knowledge, no prior work has explored learning a camera-agnostic post-illuminant-estimation mapping in a canonical space specifically aimed at cross-camera AWB with white-balance preferences.

\section{Method}
\label{sec:method}

\begin{figure}[t]
\centering
\includegraphics[width=\linewidth]{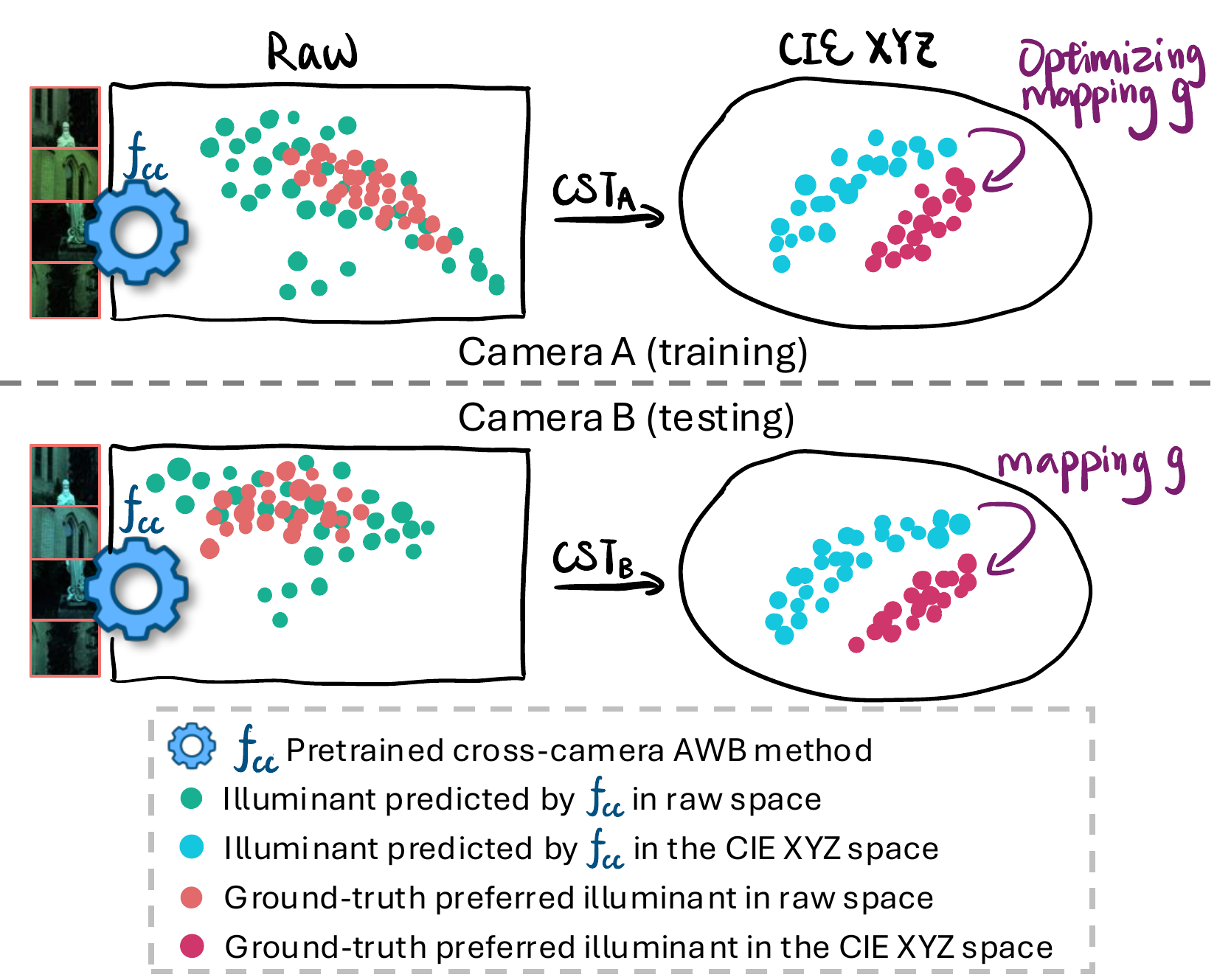}
\vspace{-4mm}
\caption{
Our method learns a camera-agnostic mapping from illuminants predicted by a cross-camera neutral AWB algorithm to a target white-balance preference in the CIE XYZ space. At test time, it acts as a plug-and-play module that maps neutral illuminants to the desired preference for unseen cameras---without retraining or tuning. The mapped illuminant is then transformed back to the unseen camera’s raw RGB space.
\label{fig:main}}
\vspace{-2mm}
\end{figure}

Given an RGB illuminant color, $l_{\texttt{raw}} \in \mathbb{R}^{3}$, in the testing camera's raw space---estimated by a cross-camera \textit{neutral} AWB method---our goal is to estimate a mapped RGB illuminant, $\hat{l}_{\texttt{raw}} \in \mathbb{R}^{3}$, in the same raw space, aligned with a target white-balance preference that may incorporate aesthetic intent or a deliberate bias away from neutral white balance. We aim to learn a unified mapping regardless of the testing camera's raw space. In other words, the mapping should be camera-agnostic to ensure practicality and align with the goals of cross-camera AWB---eliminating the need for tuning or retraining for new cameras.


An overview of our method is shown in Fig.~\ref{fig:main}. As illustrated, our approach is learning-based and requires access to a paired training dataset from a source camera, consisting of raw images and their corresponding ground-truth white-balance preference illuminants in that camera’s raw space. These target illuminants can be obtained by an expert photographer or derived through systematic criteria with manual review. Importantly, this dataset is needed only once---we do not require retraining our model for new, unseen cameras. To achieve this and ensure practicality, our method learns a mapping from neutral illuminants---produced by any cross-camera AWB algorithm---to the corresponding white-balance preference bias, in a camera-agnostic space. Specifically, we first apply a cross-camera AWB method, $f_{cc}$, to estimate neutral illuminants for the source raw images. Both the estimated neutral illuminants and the ground-truth preferred illuminants are then transformed from the source camera’s raw space into the CIE XYZ space (Sec.~\ref{sec:raw2xyz}). We then learn a mapping function, $g$, that translates neutral illuminants to their white-balance-preferred counterparts in this common space (Sec.~\ref{sec:mapping}).

At inference time, this learned mapping can be directly applied to raw images from new, unseen cameras. That is, we estimate the neutral illuminant in the testing camera's raw space using a cross-camera AWB method, transform it to the CIE XYZ space, apply the learned mapping, and finally convert the result back to the testing camera’s raw space. Our overall framework can be summarized as:

\begin{equation}
\label{eq:main}
l_{\texttt{raw}} = f_{cc}(\mathbf{I_{\texttt{raw}}}),
\end{equation}

\begin{equation}
\hat{l}_{\texttt{raw}} = \texttt{CST}^{-1}_{l_{\texttt{raw}}} g\left(\texttt{CST}_{l_{\texttt{raw}}} l_{\texttt{raw}}\right),
\end{equation}

\noindent where $\mathbf{I_{\texttt{raw}}}$ is the downsampled raw image (which may optionally include additional images or metadata required by the cross-camera AWB method, $f_{cc}$), and $l_{\texttt{raw}} \in \mathbb{R}^{3}$ is the illuminant in raw space predicted by a cross-camera AWB method---either a pretrained learning-based model or a statistical method that is inherently cross-camera. $\texttt{CST}_{l_{\texttt{raw}}}$ denotes the $3\!\times\!3$ color space transformation matrix computed based on $l_{\texttt{raw}}$. The function $g$ represents the learned mapping applied in the CIE XYZ space.

\subsection{Raw-to-XYZ Conversion}
\label{sec:raw2xyz}
The CIE XYZ color space is a standardized, device-independent color space defined by the International Commission on Illumination (CIE). It was designed to model human color perception and serves as a foundational reference for many other color spaces (e.g., sRGB, LAB, etc.). In our work, we map estimated neutral illuminants to the CIE XYZ space before applying our mapping. The motivation for choosing the CIE XYZ space is that camera manufacturers typically pre-calibrate CST matrices for each new camera as part of the manufacturing process. These matrices are designed to map the camera’s raw RGB space---under a specific correlated color temperature (CCT)---to the CIE XYZ space. They are accessible to camera ISPs and are often stored in the metadata of raw files. In the case of DNG files---a standardized raw format---these matrices are stored in the `ColorMatrix1' and `ColorMatrix2' tags, each calibrated under a standard illuminant specified by the `CalibrationIlluminant1' and `CalibrationIlluminant2' tags, respectively.

To compute a CST matrix for an arbitrary illuminant (different from the two calibration illuminants), we follow the DNG specification \cite{dng} and perform linear interpolation based on the inverse CCT of the target illuminant \cite{graphics2raw}:

\begin{equation}
\label{eq:cst}
\texttt{CST}_{l_{\texttt{raw}}} = \alpha\left(l_{\texttt{raw}}\right)\texttt{CST}_{l_a} + \left(1-\alpha\left(l_{\texttt{raw}}\right)\right)\texttt{CST}_{l_b},
\end{equation}
\begin{equation}
\label{eq:cct}
\alpha\left(l_{\texttt{raw}}\right) = \frac{1/c_{l_{\texttt{raw}}} - 1/c_{l_b}}{1/c_{l_a} - 1/c_{l_b}},
\end{equation}

\noindent where $\texttt{CST}_{l_a}$ and $\texttt{CST}_{l_b}$ are the CST matrices pre-calibrated for standard illuminants $l_a$ and $l_b$, and $c_l$ denotes the CCT of illuminant $l$, computed using Robertson’s method \cite{cct1968robertson}.

We use this transformation to convert all training illuminants (both neutral and preferred) from the training camera's raw space to the CIE XYZ space. At inference time, we apply the same transformation to the neutral illuminant estimated by the cross-camera AWB method, $f_{cc}$, converting it from the testing camera's raw space to the CIE XYZ space. After our learned mapping (Sec.~\ref{sec:mapping}), we apply the inverse of the computed CST matrix (Eq.~\ref{eq:cst}) to transform the mapped illuminant back from the CIE XYZ space to the testing camera’s raw space. This design enables our mapping to be camera-agnostic, allowing it to serve as a plug-and-play module for any unseen camera without requiring retraining or tuning.

\subsection{Learnable White-Balance Preference Mapping}
\label{sec:mapping}
We design $g$ as a learnable function implemented by a lightweight neural network, trained to map neutral illuminant colors in the CIE XYZ space to their corresponding target white-balance preferences. Our model consists of 539 parameters and comprises four linear layers with ELU activations \cite{clevert2015fast}, with batch normalization applied to the first layer. To enrich the input feature, we first project the input neutral illuminant color, $l_{\texttt{xyz}} = \texttt{CST}_{l_{\texttt{raw}}}l_{\texttt{raw}}$, $l_{\texttt{xyz}} \in \mathbb{R}^{3}$, to a higher-dimensional space and process it using our network, $g$, as follows:
\begin{equation}
\hat{l}_{\texttt{xyz}}= g(\phi(l_{\texttt{xyz}})),\end{equation}
\noindent where $\phi$ is a polynomial kernel \cite{hong2001study}, which transforms the [$x$, $y$, $z$]$^T$ illuminant vector into [$x$, $y$, $z$, $xy$, $xz$, $yz$, $x^2$, $y^2$, $z^2$, $xyz$]$^T$.
Each layer of our network has the following (input, output) dimensions: (10, 16), (16, 8), (8, 16), and (16, 3), respectively.
The last layer outputs the predicted illuminant, $\hat{l}_{\texttt{xyz}}$, and has no activation. $\hat{l}_{\texttt{xyz}}$ is then normalized via division by its L2 norm before transforming back to the target sensor's raw space.

The weights of our mapping function, $g$, are optimized using the Adam optimizer \cite{adam} for 2000 epochs, with $\beta$ values set to (0.9, 0.999), a weight decay of $10^{-9}$, and a cosine annealing learning rate schedule \cite{cosineAnnealing}. The objective is to minimize the angular error between the mapped illuminants (in CIE XYZ) and the ground-truth illuminants with the target white-balance preference in the training dataset.

\section{Experiments}
\label{sec:experiments}
We evaluate our method in a cross-camera setting, where the model is trained on data captured by a source camera and tested on data from three different target cameras to assess generalization. To validate performance in a camera-specific scenario, we also report results on the same camera used for training. In addition, we compare our approach against two baseline mappings and analyze its performance when the mapping is learned in an alternative color space. We follow standard evaluation practices in color constancy and white balance literature, reporting angular error statistics including the mean, median, and tri-mean, as well as the arithmetic mean of the best 25\%, worst 25\%, and worst 5\% angular errors between the predicted and ground-truth illuminants, in the target camera's raw space.

\subsection{Data}
\label{sec:dataset}
To quantitatively evaluate our method, we require paired data captured by different cameras, each with distinct characteristics, where a ground-truth aesthetically-preferred illuminant accompanies each image. To ensure consistency in the aesthetic adjustments of the ground-truth illuminants across all cameras, we use the in-camera AWB estimated illuminant values saved in the raw image metadata. We use these values as our target white-balance preference because AWB algorithms from the same manufacturer, particularly across successive smartphone generations, are typically tuned to maintain a consistent aesthetic style. 

In our experiments, we use the Samsung S24 Ultra’s wide camera (S24U-W) for training, utilizing the S24 dataset introduced in \cite{afifi2025time}, which includes 2,619 training images, 205 validation images, and 400 testing raw images captured under diverse lighting conditions. This dataset also provides aesthetically-preferred white-balance annotations, where an expert photographer has manually adjusted the white balance of each image.

In the cross-camera setup, we use the in-camera AWB estimated illuminants as the target white-balance preference to evaluate generalization. Additionally, we report results using the aesthetically-preferred white-balance annotations in the S24 dataset \cite{afifi2025time} in the camera-specific evaluation.  For cross-camera testing, we collected 257 additional scenes using three cameras from the same manufacturer as the training device. Specifically, the testing images were captured using the Samsung S25 Ultra’s wide, telephoto, and super-telephoto cameras---referred to as S25U-W, S25U-T, and S25U-ST, respectively---resulting in a total of 771 raw images.

\subsection{Cross-Camera Experiments}
\label{sec:cross-cam-expt}
\begin{table}[!t]
\centering
\caption{\textbf{Cross-camera evaluation:} We report the angular errors on the \textit{S25U-W} test camera for different cross-camera methods combined with  mapping baselines and our proposed mapping.
}
\vspace{-2mm}
\scalebox{0.65}{
\begin{tabular}{c|l|ccccccc}
\begin{tabular}[c]{@{}c@{}c@{}}\textbf{Cross} \\ \textbf{camera} \\ \textbf{method}\end{tabular} & \textbf{Mapping} & \textbf{Mean}  & \textbf{Med.}  & \begin{tabular}[c]{@{}c@{}}\textbf{Best} \\ \textbf{25\%}\end{tabular} & \begin{tabular}[c]{@{}c@{}}\textbf{Worst} \\ \textbf{25\%}\end{tabular} & \begin{tabular}[c]{@{}c@{}}\textbf{Worst} \\ \textbf{5\%}\end{tabular} & \textbf{Tri.}   & \textbf{Max} \\ \hline\hline
\multirow{4}{*}{C5} & No mapping & 2.77 & 1.34 & 0.43 & 7.61 & 19.96 & 1.52 & 41.22 \\
                    & $3 \times 3$ & 2.53 & 1.91 & 0.77 & 5.29 & 11.91 & 2.01 & 26.78 \\
                    & Polynomial & 2.21 & 1.10 & 0.36 & 6.03 & 17.34 & 1.24 & 45.75 \\
                    & Ours & \textbf{1.34} & \textbf{0.92} & \textbf{0.30} & \textbf{3.07} & \textbf{5.66} & \textbf{1.02} & \textbf{9.07} \\ \hline
\multirow{4}{*}{C4} & No mapping & 3.33 & 2.06 & 0.61 & 8.24 & 23.38 & 2.24 & 48.11 \\
                    & $3 \times 3$, & 3.49 & 2.78 & 1.43 & 6.49 & 9.44 & 3.06 & 17.59 \\
                    & Polynomial & 2.51 & 1.88 & \textbf{0.49} & 5.88 & 14.82 & 1.88 & 44.57 \\
                    & Ours & \textbf{1.95} & \textbf{1.71} & 0.68 & \textbf{3.63} & \textbf{5.51} & \textbf{1.75} & \textbf{8.81} \\ \hline
\multirow{4}{*}{GW} & No mapping & 3.60 & 2.43 & 0.63 & 9.02 & 24.25 & 2.40 & 49.89 \\
                    & $3 \times 3$ & 2.74 & 1.86 & 0.72 & 6.11 & 12.74 & 2.16 & 33.70 \\
                    & Polynomial & 2.27 & 1.14 & 0.44 & 6.32 & 21.47 & 1.21 & 72.45 \\
                    & Ours & \textbf{1.14} & \textbf{0.93} & \textbf{0.35} & \textbf{2.34} & \textbf{4.18} & \textbf{0.94} & \textbf{8.10} \\ \hline
\multirow{4}{*}{wGE} & No mapping & 1.83 & \textbf{0.89} & \textbf{0.25} & 5.13 & 11.24 & \textbf{1.04} & 18.74 \\
                    & $3 \times 3$ & 2.80 & 2.23 & 0.99 & 5.51 & 7.63 & 2.45 & \textbf{10.33} \\
                    & Polynomial & 1.79 & 1.21 & 0.43 & 4.13 & 8.36 & 1.29 & 17.01 \\
                    & Ours & \textbf{1.59} & 1.09 & 0.41 & \textbf{3.66} & \textbf{7.34} & 1.15 & 11.72 \\

\end{tabular}
}
\label{tab:results-s25u-w}
\end{table}
\begin{table}[!t]
\centering
\caption{\textbf{Cross-camera evaluation:} We report the angular errors on the \textit{S25U-T} test camera for different cross-camera methods combined with mapping baselines and our proposed mapping.
}
\vspace{-2mm}
\scalebox{0.65}{
\begin{tabular}{c|l|ccccccc}
\begin{tabular}[c]
{@{}c@{}c@{}}\textbf{Cross} \\ \textbf{camera} \\ \textbf{method}\end{tabular} & \textbf{Mapping} & \textbf{Mean}  & \textbf{Med.}  & \begin{tabular}[c]{@{}c@{}}\textbf{Best} \\ \textbf{25\%}\end{tabular} & \begin{tabular}[c]{@{}c@{}}\textbf{Worst} \\ \textbf{25\%}\end{tabular} & \begin{tabular}[c]{@{}c@{}}\textbf{Worst} \\ \textbf{5\%}\end{tabular} & \textbf{Tri.}   & \textbf{Max} \\ \hline\hline
\multirow{4}{*}{C5} & No mapping & 3.90 & 3.06 & 0.97 & 8.17 & 17.41 & 3.18 & 34.77 \\
                    & $3 \times 3$ & 2.47 & 1.79 & 0.57 & 5.61 & 12.26 & 1.90 & 23.90 \\ 
                    & Polynomial & 2.26 & 1.70 & \textbf{0.53} & 5.02 & 9.83 & 1.78 & 15.90 \\ 
                    & Ours & \textbf{2.10} & \textbf{1.61} & 0.58 & \textbf{4.44} & \textbf{7.60} & \textbf{1.71} & \textbf{10.14} \\ \hline 
\multirow{4}{*}{C4} & No mapping & 4.05 & 2.81 & 0.75 & 9.74 & 24.89 & 2.84 & 49.18 \\
                    & $3 \times 3$ & 4.00 & 3.16 & 1.41 & 7.97 & 14.15 & 3.39 & 30.72 \\
                    & Polynomial & 2.60 & 2.03 & 0.66 & 5.64 & 11.01 & 2.10 & 20.03 \\
                    & Ours & \textbf{2.17} & \textbf{1.78} & \textbf{0.56} & \textbf{4.51} & \textbf{7.43} & \textbf{1.84} & \textbf{10.84} \\ \hline 
\multirow{4}{*}{GW} & No mapping & 4.53 & 3.18 & 0.97 & 10.57 & 25.11 & 3.38 & 47.37 \\
                    & $3 \times 3$ & 3.23 & 2.32 & 1.12 & 6.84 & 15.57 & 2.47 & 39.54 \\
                    & Polynomial & 3.15 & 1.56 & 0.54 & 8.62 & 26.97 & 1.70 & 82.51 \\
                    & Ours & \textbf{1.77} & \textbf{1.31} & \textbf{0.52} & \textbf{3.84} & \textbf{6.64} & \textbf{1.40} & \textbf{14.48} \\ \hline 
\multirow{4}{*}{wGE} & No mapping & 3.60 & 1.77 & \textbf{0.41} & 10.27 & 24.74 & 2.08 & 44.09 \\
                    & $3 \times 3$ & 3.63 & 2.69 & 1.12 & 7.71 & 15.75 & 2.87 & 34.68 \\
                    & Polynomial & 3.44 & 1.71 & 0.67 & 9.49 & 25.02 & 1.92 & 64.48 \\
                    & Ours & \textbf{2.20} & \textbf{1.29} & 0.48 & \textbf{5.50} & \textbf{9.73} & \textbf{1.48} & \textbf{13.03} \\
\end{tabular}
}
\label{tab:results-s25u-t}
\end{table}
\begin{table}[!t]
\centering
\caption{\textbf{Cross-camera evaluation:} We report the angular errors on the \textit{S25U-ST} test camera for different cross-camera methods combined with mapping baselines and our proposed mapping.
}
\scalebox{0.65}{
\begin{tabular}{c|l|ccccccc}
\begin{tabular}[c]{@{}c@{}c@{}}\textbf{Cross} \\ \textbf{camera} \\ \textbf{method}\end{tabular} & \textbf{Mapping} & \textbf{Mean}  & \textbf{Med.}  & \begin{tabular}[c]{@{}c@{}}\textbf{Best} \\ \textbf{25\%}\end{tabular} & \begin{tabular}[c]{@{}c@{}}\textbf{Worst} \\ \textbf{25\%}\end{tabular} & \begin{tabular}[c]{@{}c@{}}\textbf{Worst} \\ \textbf{5\%}\end{tabular} & \textbf{Tri.}   & \textbf{Max} \\ \hline\hline
    \multirow{4}{*}{C5} & No mapping & 4.18 & 3.41 & 1.04 & 8.89 & 17.58 & 3.47 & 33.92 \\
                        & $3 \times 3$ & 2.53 & \textbf{1.82} & \textbf{0.58} & 5.68 & 11.93 & \textbf{1.94} & 22.07 \\
                        & Polynomial & 2.49 & 1.96 & 0.61 & 5.28 & 9.15 & 2.04 & 11.82 \\
                        & Ours & \textbf{2.45} & 1.98 & 0.69 & \textbf{5.14} & \textbf{8.50} & 2.03 & \textbf{11.00} \\ \hline
    \multirow{4}{*}{C4} & No mapping & 4.20 & 2.75 & 0.82 & 10.05 & 22.11 & 3.01 & 43.68 \\
                        & $3 \times 3$ & 3.68 & 2.81 & 1.23 & 7.38 & 12.12 & 3.11 & 20.18 \\
                        & Polynomial & \textbf{2.67} & \textbf{1.98} & \textbf{0.56} & 6.05 & 11.48 & \textbf{2.11} & 24.50 \\
                        & Ours & 2.69 & 2.20 & 0.85 & \textbf{5.45} & \textbf{9.67} & 2.27 & \textbf{13.90} \\ \hline
    \multirow{4}{*}{GW} & No mapping & 4.87 & 3.37 & 0.94 & 11.53 & 24.34 & 3.60 & 43.99 \\
                        & $3 \times 3$ & 2.86 & 2.07 & 0.72 & 6.38 & 13.89 & 2.20 & 31.41 \\
                        & Polynomial & 2.16 & 1.50 & \textbf{0.54} & 5.00 & 10.71 & 1.58 & 15.98 \\
                        & Ours & \textbf{1.89} & \textbf{1.42} & 0.55 & \textbf{4.06} & \textbf{7.60} & \textbf{1.48} & \textbf{10.40} \\  \hline
    \multirow{4}{*}{wGE} & No mapping & 4.32 & 2.00 & \textbf{0.56} & 12.23 & 27.86 & 2.33 & 51.39 \\
                        & $3 \times 3$ & 3.54 & 2.47 & 0.79 & 8.17 & 17.56 & 2.64 & 43.53 \\
                        & Polynomial & 2.98 & 1.77 & 0.58 & 7.69 & 15.10 & 1.96 & 26.30 \\
                        & Ours & \textbf{2.55} & \textbf{1.67} & 0.64 & \textbf{6.06} & \textbf{10.46} & \textbf{1.82} & \textbf{15.30} \\
\end{tabular}
}
\label{tab:results-s25u-st}
\end{table}
\begin{table*}[!t]
\centering
\caption{\textbf{Camera-specific evaluation:} Angular errors on the test set of the \textit{S24U-W} camera \cite{afifi2025time}, which is also the same camera used for training our model and fitting the baseline mappings. Results are reported for both the in-camera AWB estimated illuminant ground truth and the aesthetically-preferred ground truth provided in the S24 dataset, shown in the order: camera-estimated / aesthetically-preferred.}
\vspace{-2mm}
\scalebox{0.68}{
\begin{tabular}{c|l|ccccccc}
\begin{tabular}[c]{@{}c@{}}\textbf{Cross-camera} \\ \textbf{method}\end{tabular} & \textbf{Mapping} & \textbf{Mean}  & \textbf{Med.}  & \begin{tabular}[c]{@{}c@{}}\textbf{Best} \\ \textbf{25\%}\end{tabular} & \begin{tabular}[c]{@{}c@{}}\textbf{Worst} \\ \textbf{25\%}\end{tabular} & \begin{tabular}[c]{@{}c@{}}\textbf{Worst} \\ \textbf{5\%}\end{tabular} & \textbf{Tri.}   & \textbf{Max} \\ \hline\hline
\multirow{4}{*}{C5} & No Mapping & 2.92 / 2.96 & 1.71 / 1.93 & 0.50 / 0.57 & 7.34 / 7.22 & 14.71 / 14.18 & 2.03 / 2.14 & 35.44 / 28.35 \\
                    & $3 \times 3$ & 2.57 / 2.33 & 2.26 / 2.01 & 0.90 / 0.79 & 4.83 / 4.46 & 8.77 / 7.94 & 2.29 / 2.04 & 24.83 / 16.72 \\
                    & Ours & \textbf{1.44} / \textbf{1.56} & \textbf{0.98} / \textbf{1.10} & \textbf{0.30} / \textbf{0.32} & \textbf{3.31} / \textbf{3.61} & \textbf{5.81} / \textbf{6.53} & \textbf{1.10} / \textbf{1.19} & \textbf{13.06} / \textbf{12.90} \\ \hline
\multirow{4}{*}{C4} & No Mapping & 3.67 / 3.01 & 2.63 / 1.88 & 0.81 / 0.61 & 8.46 / 7.21 & 15.77 / 14.03 & 2.91 / 2.21 & 37.53 / 34.68 \\
                    & $3 \times 3$ & 4.24 / 3.89 & 3.84 / 3.54 & 1.89 / 1.98 & 7.28 / 6.42 & 10.14 / 8.74 & 3.91 / 3.63 & 14.90 / 12.69 \\
                    & Ours & \textbf{1.96} / \textbf{1.73} & \textbf{1.47} / \textbf{1.41} & \textbf{0.57} / \textbf{0.39} & \textbf{4.17} / \textbf{3.64} & \textbf{6.62} / \textbf{5.93} & \textbf{1.62} / \textbf{1.48} & \textbf{11.66} / \textbf{10.38} \\ \hline
\multirow{4}{*}{GW} & No Mapping & 4.72 / 5.63 & 3.38 / 4.65 & 0.98 / 1.18 & 10.65 / 11.89 & 19.40 / 19.69 & 3.70 / 4.79 & 36.01 / 31.63 \\
                    & $3 \times 3$ & 2.96 / 2.87 & 2.60 / 2.52 & 1.19 / 0.98 & 5.41 / 5.35 & 9.11 / 8.55 & 2.63 / 2.58 & 24.55 / 18.14 \\
                    & Ours & \textbf{1.73} / \textbf{1.97} & \textbf{1.24} / \textbf{1.43} & \textbf{0.37} / \textbf{0.36} & \textbf{3.98} / \textbf{4.52} & \textbf{6.36} / \textbf{7.10} & \textbf{1.36} / \textbf{1.54} & \textbf{8.47} / \textbf{9.64} \\ \hline
\multirow{4}{*}{wGE} & No Mapping & 3.15 / 3.33 & 1.82 / 2.01 & 0.47 / 0.53 & 8.19 / 8.45 & 16.94 / 16.01 & 2.03 / 2.33 & 37.62 / 31.94 \\
                    & $3 \times 3$ & 3.28 / 2.95 & 2.66 / 2.47 & 1.23 / 1.09 & 6.42 / 5.65 & 10.85 / 9.02 & 2.80 / 2.58 & 29.95 / 22.73 \\
                    & Ours & \textbf{1.70} / \textbf{1.79} & \textbf{1.28} / \textbf{1.17} & \textbf{0.37} / \textbf{0.30} & \textbf{3.82} / \textbf{4.35} & \textbf{6.83} / \textbf{7.91} & \textbf{1.37} / \textbf{1.32} & \textbf{10.80} / \textbf{17.49} \\

\end{tabular}
}
\label{tab:results-s24u-w}
\end{table*}
\begin{table*}[!t]
\centering
\caption{\textbf{Ablation study:} We report results for different cross-camera methods using our learnable mapping, optimized in (1) the training camera's raw space and (2) the CIE XYZ space, applied across three unseen test cameras: \textit{S25U-W}, \textit{S25U-T}, and \textit{S25U-ST}.
}
\vspace{-2mm}
\scalebox{0.7}{
\begin{tabular}{c|c|c|ccccccc}
\textbf{Camera} & \begin{tabular}[c]{@{}c@{}}\textbf{Cross-camera} \\ \textbf{method}\end{tabular} & \textbf{Mapping space} & \textbf{Mean}  & \textbf{Med.}  & \begin{tabular}[c]{@{}c@{}}\textbf{Best} \\ \textbf{25\%}\end{tabular} & \begin{tabular}[c]{@{}c@{}}\textbf{Worst} \\ \textbf{25\%}\end{tabular} & \begin{tabular}[c]{@{}c@{}}\textbf{Worst} \\ \textbf{5\%}\end{tabular} & \textbf{Tri.}   & \textbf{Max} \\ \hline\hline

\multirow{8}{*}{S25U-W} & \multirow{2}{*}{C5} & Camera's raw & 1.67 & 1.26 & 0.48 & 3.52 & 6.00 & 1.34 & 9.25 \\
                        &                     & CIE XYZ (ours) & \textbf{1.34} & \textbf{0.92} & \textbf{0.30} & \textbf{3.07} & \textbf{5.66} & \textbf{1.02} & \textbf{9.07} \\ \cline{2-10} 
                        & \multirow{2}{*}{C4} & Camera's raw & 2.89 & 2.90 & 1.12 & 4.79 & 6.93 & 2.78 & 10.58 \\
                        &                    & CIE XYZ (ours) & \textbf{1.95} & \textbf{1.71} & \textbf{0.68} & \textbf{3.63} & \textbf{5.51} & \textbf{1.75} & \textbf{8.81} \\ \cline{2-10}
                        & \multirow{2}{*}{GW} & Camera's raw & 2.10 & 2.03 & 0.85 & 3.52 & 5.05 & 2.00 & \textbf{7.66} \\
                        &                    & CIE XYZ (ours) & \textbf{1.14} & \textbf{0.93} & \textbf{0.35} & \textbf{2.34} & \textbf{4.18} & \textbf{0.94} & 8.10 \\ \cline{2-10}
                        & \multirow{2}{*}{wGE} & Camera's raw & 1.95 & 1.75 & 0.65 & 3.75 & \textbf{6.71} & 1.72 & \textbf{11.10} \\
                        &                    & CIE XYZ (ours) & \textbf{1.59} & \textbf{1.09} & \textbf{0.41} & \textbf{3.66} & 7.34 & \textbf{1.15} & 11.72 \\ \hline

\multirow{8}{*}{S25U-T} & \multirow{2}{*}{C5} & Camera's raw & 2.13 & 1.65 & 0.59 & 4.57 & 8.60 & 1.71 & 11.41 \\
                        &                    & CIE XYZ (ours) & \textbf{2.10} & \textbf{1.61} & \textbf{0.58} & \textbf{4.44} & \textbf{7.60} & \textbf{1.71} & \textbf{10.14} \\ \cline{2-10}
                        & \multirow{2}{*}{C4} & Camera's raw & 3.49 & 3.22 & 1.38 & 6.11 & 9.16 & 3.24 & 12.56 \\
                        &                    & CIE XYZ (ours) & \textbf{2.17} & \textbf{1.78} & \textbf{0.56} & \textbf{4.51} & \textbf{7.43} & \textbf{1.84} & \textbf{10.84} \\ \cline{2-10}
                        & \multirow{2}{*}{GW} & Camera's raw & 2.18 & 1.95 & 0.85 & 4.01 & 6.78 & 1.94 & \textbf{13.93} \\
                        &                    & CIE XYZ (ours) & \textbf{1.77} & \textbf{1.31} & \textbf{0.52} & \textbf{3.84} & \textbf{6.64} & \textbf{1.40} & 14.48 \\ \cline{2-10}
                        & \multirow{2}{*}{wGE}& Camera's raw & 2.41 & 1.87 & 0.68 & \textbf{5.19} & \textbf{8.72} & 1.95 & \textbf{12.80} \\
                        &                    & CIE XYZ (ours) & \textbf{2.20} & \textbf{1.29} & \textbf{0.48} & 5.50 & 9.73 & \textbf{1.48} & 13.03 \\ \hline

\multirow{8}{*}{S25U-ST} & \multirow{2}{*}{C5} & Camera's raw & 4.06 & 3.79 & 1.55 & 7.20 & 11.27 & 3.78 & 15.04 \\
                        &                    & CIE XYZ (ours) & \textbf{2.45} & \textbf{1.98} & \textbf{0.69} & \textbf{5.14} & \textbf{8.50} & \textbf{2.03} & \textbf{11.00} \\ \cline{2-10}
                        & \multirow{2}{*}{C4} & Camera's raw & 5.34 & 5.17 & 2.78 & 8.24 & 12.04 & 5.14 & 17.77 \\
                        &                    & CIE XYZ (ours) & \textbf{2.69} & \textbf{2.20} & \textbf{0.85} & \textbf{5.45} & \textbf{9.67} & \textbf{2.27} & \textbf{13.90} \\ \cline{2-10}
                        & \multirow{2}{*}{GW} & Camera's raw & 4.33 & 4.14 & 2.17 & 6.90 & 9.79 & 4.11 & 12.40 \\
                        &                    & CIE XYZ (ours) & \textbf{1.89} & \textbf{1.42} & \textbf{0.55} & \textbf{4.06} & \textbf{7.60} & \textbf{1.48} & \textbf{10.40} \\ \cline{2-10} 
                        & \multirow{2}{*}{wGE} & Camera's raw & 4.29 & 4.03 & 1.88 & 7.34 & 11.67 & 3.96 & 19.86 \\
                        &                    & CIE XYZ (ours) & \textbf{2.55} & \textbf{1.67} & \textbf{0.64} & \textbf{6.06} & \textbf{10.46} & \textbf{1.82} & \textbf{15.30} \\

\end{tabular}
}
\label{tab:results-raw-vs-xyz}
\end{table*}

\begin{figure*}[t]
\centering
\includegraphics[width=0.94\linewidth]{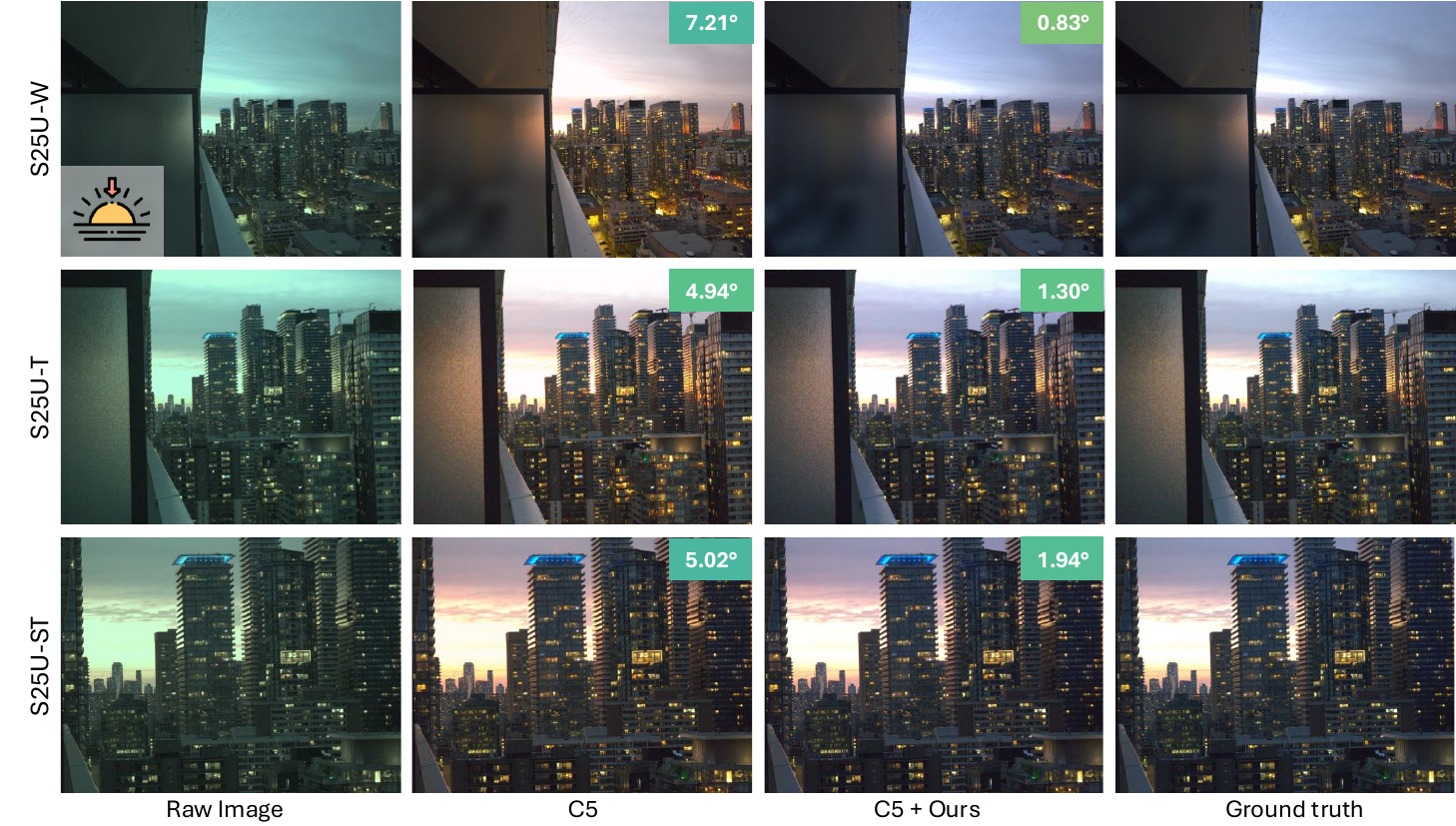}
\vspace{-1mm}
\caption{Qualitative results. Shown is a sunset scene captured by the S25U wide, telephoto, and super-telephoto sensors, white-balanced using C5 \cite{C5} predicted illuminant, C5 corrected with our mapping function, and the ground truth aesthetically-preferred illuminant. \label{fig:qualitative-c5}}
\end{figure*}

\begin{figure*}[t]
\centering
\includegraphics[width=0.94\linewidth]{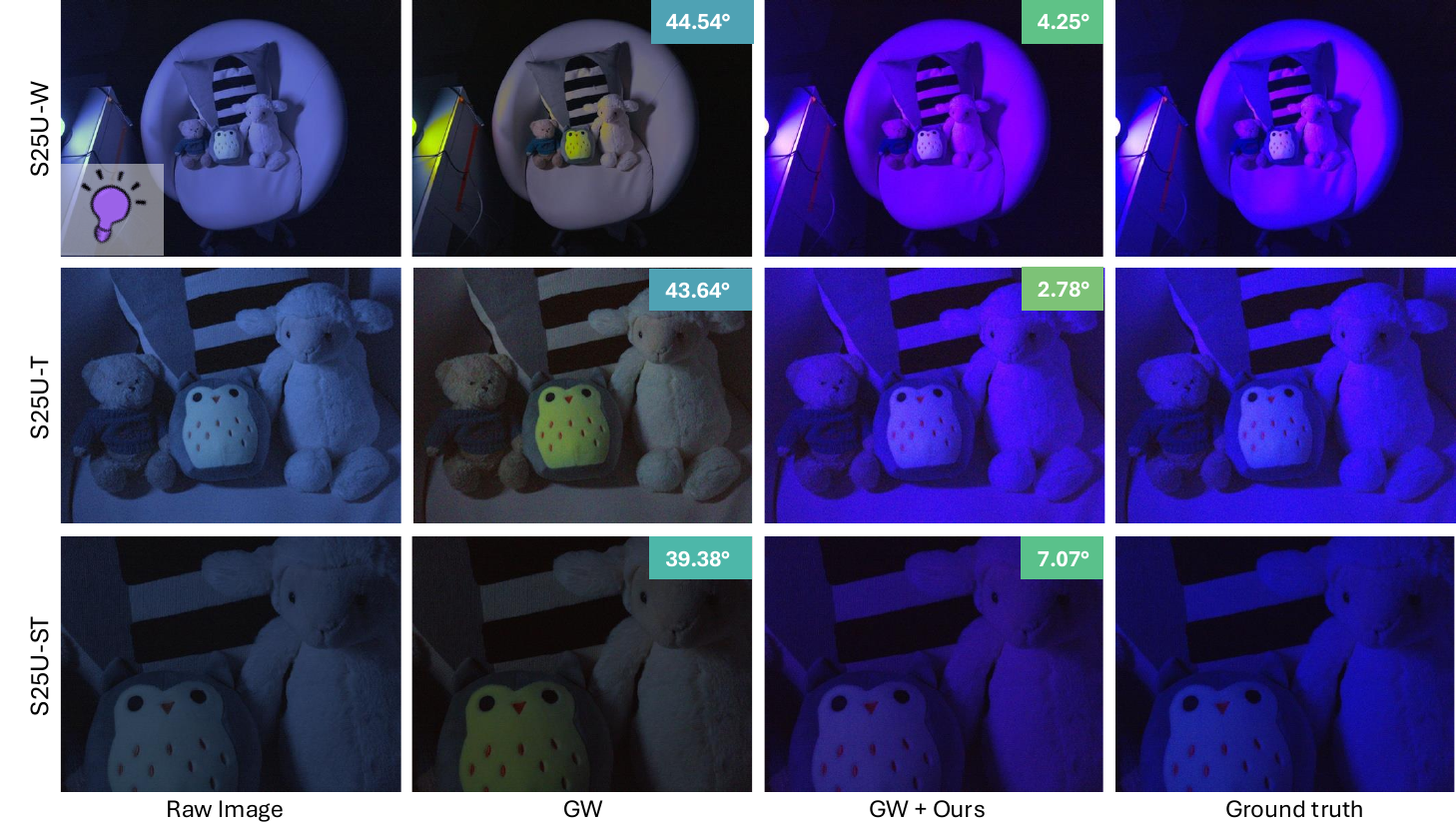}
\vspace{-1mm}
\caption{Qualitative results. Shown is an indoor scene illuminated by a purple LED light, captured by the S25U wide, telephoto, and super-telephoto sensors, white-balanced using GW \cite{GW} predicted illuminant, GW corrected with our mapping function, and the ground truth aesthetically-preferred illuminant. \label{fig:qualitative-gw}}
\end{figure*}

We report the results of our model, trained on S24U-W, on the S25U-W test set in Table~\ref{tab:results-s25u-w}, the S25U-T test set in Table~\ref{tab:results-s25u-t}, and the S25U-ST test set in Table~\ref{tab:results-s25u-st}. We show results basing our method on four cross-camera AWB methods, including two learning-based methods, C5 \cite{C5} and C4 \cite{C4}, and two statistical-based methods, gray-world (GW) \cite{GW} and weighted gray-edge (wGE) \cite{wGE}. Qualitative results are shown in Figs.~\ref{fig:qualitative-c5}--\ref{fig:qualitative-gw}, with additional examples in the supplemental materials.

We compare our method against three mapping baselines: (1) no mapping, (2) polynomial mapping \cite{polymapping}, and (3) $3\!\times\!3$ correction \cite{MomentCorrection}. For both the polynomial mapping and the $3\!\times\!3$ correction, we use the S24 dataset \cite{afifi2025time}---consistent with our method---to fit the coefficients of the respective mapping functions. In the `no mapping' baseline, we compute the angular error between the cross-camera method's output and the white-balance preferred ground-truth illuminant. As expected, the angular error is relatively high, since cross-camera methods are designed to predict neutral rather than stylistically biased illuminants. The $3\!\times\!3$ correction baseline also yields high errors, sometimes even higher than the `no mapping' baseline, likely due to its limited expressive capacity, thus underfitting the training dataset. The polynomial mapping performs better, offering higher flexibility than the $3\!\times\!3$ correction. Our model combines high expressiveness with lightweight computation, consistently outperforming all baselines in accuracy.

Notably, our mapping performs particularly well when combined with the GW method \cite{GW}, often surpassing combinations with learning-based approaches. This may be attributed to GW's more systematic and predictable failures \cite{zakizadeh2015hybrid}, which make it easier for our learned mapping to not only bias the initial neutral illuminant estimates towards the preferred white-balance target but also correct those systematic errors. A similar trend was observed in APAP \cite{APAP}.

\vspace{5pt}\noindent\textbf{Ablation.}
We perform an ablation study on the choice of color space by learning the mapping directly in raw space instead of the camera-agnostic CIE XYZ space. As shown in Table~\ref{tab:results-raw-vs-xyz}, learning the mapping in the CIE XYZ consistently reduces angular error across all cameras and cross-camera base methods used for initial prediction. This aligns with our expectations: mappings learned in raw space tend to capture camera-specific spectral responses and characteristics (see Fig.~\ref{fig:raw_colorchart}), which limits their generalization to other cameras.

\subsection{Camera-Specific Experiments}
For completeness, we report results on the test split of the S24 dataset \cite{afifi2025time} in Table~\ref{tab:results-s24u-w}, to evaluate the effectiveness of our method for camera-specific white balance. In addition to using the camera-estimated illuminant values as ground truth, we also present results based on photographer-annotated aesthetically-preferred illuminants in the S24 dataset. Our proposed mapping consistently outperforms all baselines under both ground-truth settings.

\section{Conclusion}
\label{sec:conclusion}

We propose a cross-camera auto white balance method that explicitly accounts for target white-balance preferences. Specifically, we present a lightweight model (with only $\sim$500 parameters) that learns in a camera-agnostic space to map neutral illuminants---predicted by a pretrained cross-camera method---to illuminant colors that reflect a target white-balance aesthetic preference. Our approach is efficient, running in just 0.024 milliseconds on the Samsung S24 Ultra CPU, and is capable of supporting high frame rates on modern smartphone devices.

\clearpage
\maketitlesupplementary
\appendix

This supplementary material provides additional qualitative results, along with details about the dataset presented in the main paper.

\section{Data}
\label{supp-sec:dataset}
In the main paper, we mentioned that we use the in-camera AWB-estimated illuminant as the ground truth. Here, we provide further explanation for this choice.
To ensure the aesthetic consistency of the ground-truth illuminants across all cameras, we extract the in-camera AWB estimated illuminant from the raw metadata as our ground truth. 
This decision is based on the observation that AWB algorithms from the same manufacturer are typically calibrated to produce a consistent visual style, especially when the cameras are part of the same smartphone or belong to consecutive smartphone generations.
Two cameras from the same manufacturer capturing the same scene under identical illumination may yield different raw illuminant values due to differences in their sensor spectral responses and other camera-specific characteristics. However, once transformed into the sensor-agnostic CIE XYZ color space, these illuminants should converge to similar values. Significant discrepancies in this space would primarily reflect differences in the cameras' AWB biases rather than sensor characteristics.

We demonstrate that the cameras in our dataset exhibit consistent AWB aesthetic styles by capturing a Macbeth color checker under 12 different illuminations, with correlated color temperatures ranging from 2000K to 7500K. We capture the color checker using all four sensors involved in our training and evaluation: S24U-W, S25U-W, S25U-T, and S25U-ST.
As shown in Table~\ref{tab:colorchart-raw-xyz-ang-err}, the three S25U cameras show relatively high angular errors with respect to S24U-W (the training sensor) when compared in raw space. However, after converting the illuminants to the CIE XYZ color space, the angular errors are significantly reduced. This supports our observation that the in-camera AWB outputs are aligned in aesthetic style once sensor-specific differences are removed. 

\begin{figure}[t]
\centering
\includegraphics[width=\linewidth]{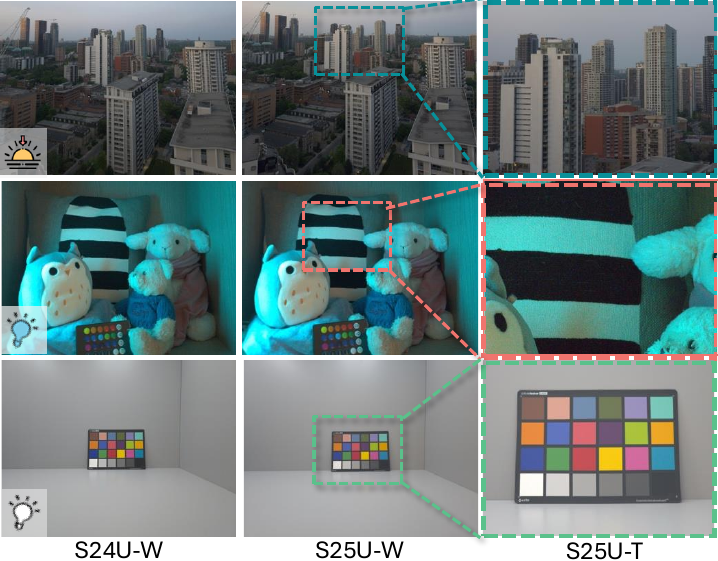}
\vspace{-4mm}
\caption{Scenes captured by different cameras from the same manufacturer (Samsung Galaxy series) often exhibit consistent white balance biases. The images are visualized by first white-balancing the raw using the respective camera-estimated illuminants, then converted to sRGB with no additional photofinishing. 
\label{fig:dataset}}
\vspace{-4mm}
\end{figure}

\begin{figure}[t]
\centering
\includegraphics[width=\linewidth]{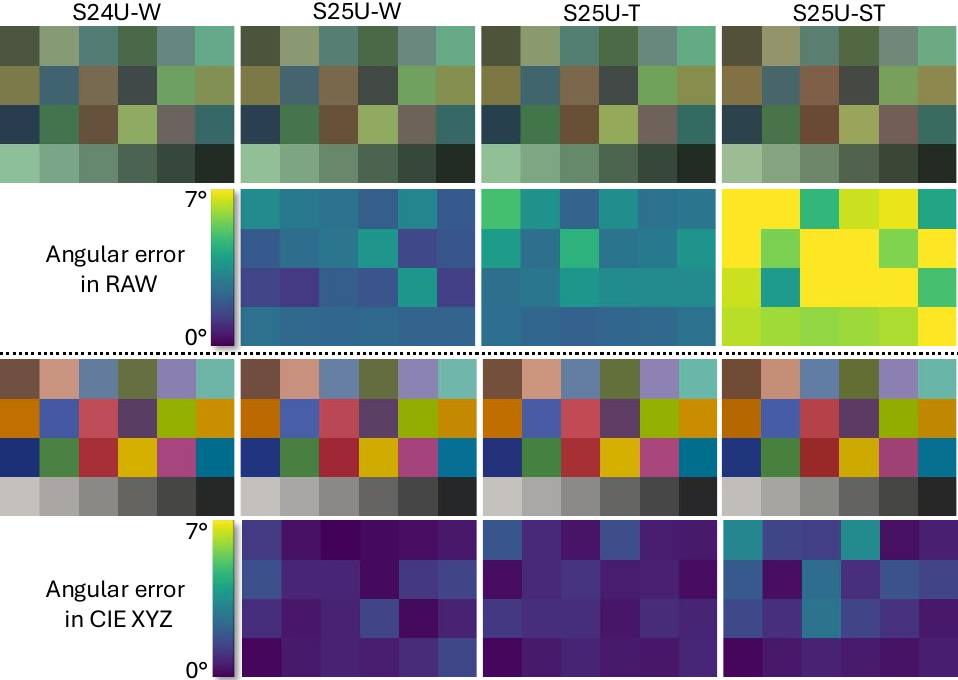}
\vspace{-4mm}
\caption{Top section: color chart captured by four cameras under the same lighting condition, shown in raw (gamma applied only to aid visualization). Angular error is computed between the three S25U cameras against S24U-W in raw. Bottom section: Angular error is computed between the three S25U cameras against S24U-W after white-balancing the color chart with the respective camera's estimated illuminant and converting to CIE XYZ (XYZ2sRGB is applied in the 3rd row, only to aid visualization).
\label{fig:dataset_colorchart}}
\vspace{-4mm}
\end{figure}

Figure~\ref{fig:dataset} further illustrates the consistency in aesthetic styles across the cameras. To remove the effects of differing fields of view between cameras, we present results using a color checker in Fig.~\ref{fig:dataset_colorchart}. We visualize the angular errors of the color checker, between S24U-W and the three S25U cameras, under the D50 illuminant in both raw (no white balance applied) and CIE XYZ (with white-balance applied under the respective camera-estimated illuminant). Angular errors reduced significantly from raw to CIE XYZ. 

\begin{figure}[t]
\centering
\includegraphics[width=\linewidth]{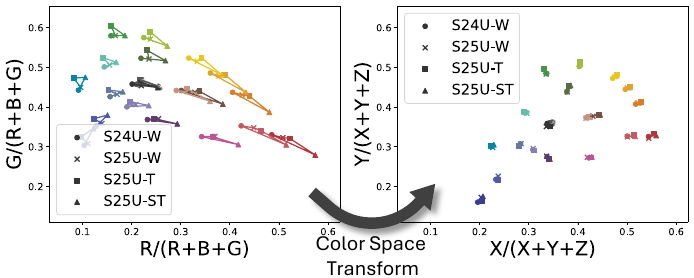}
\vspace{-4mm}
\caption{Each point in the figure represents a raw color from a Macbeth color checker. For the same object under identical illumination, different sensors produce different raw values due to variations in spectral response and other camera characteristics. After converting to CIE XYZ space, the illuminant values from different cameras converge to the same point.
\label{fig:raw_xyz_colorchart_points}}
\vspace{-4mm}
\end{figure}

The same trend can be observed from Fig.~\ref{fig:raw_xyz_colorchart_points}, another visualization of the same color chart used in Fig.~\ref{fig:dataset_colorchart}. The left side of Fig.~\ref{fig:raw_xyz_colorchart_points} shows the color values in raw RGB. This is a reproduction of Fig. 2 in the main paper, with the horizontal and vertical axes scaled to be the same as the right side of Fig.~\ref{fig:raw_xyz_colorchart_points}, which plots the color values when converted to CIE XYZ. The different colors in raw RGB converge to the same value in XYZ because the latter is a sensor-agnostic space, and the white balance algorithms onboard these smartphones are configured to produce the same aesthetic styles. 

\begin{table}[!t]
\centering
\caption{Average angular errors of S25U cameras against S24U-W in raw v.s. CIE XYZ, computed from a color chart captured under 12 illuminations ranging from 2000K to 7500K. \label{tab:colorchart-raw-xyz-ang-err}}\vspace{-1mm}
\scalebox{0.8}{
\begin{tabular}{l|ccc}

\textbf{Color space} & \textbf{S25U-W}  & \textbf{S25U-T}  & \textbf{S25U-ST} \\ \hline
Camera's raw & 2.57 & 2.51 & 6.13 \\  
CIE XYZ & 0.66 & 0.24 & 0.53 \\
\end{tabular}
}
\end{table}

\begin{figure*}[t]
\centering
\includegraphics[width=\linewidth]{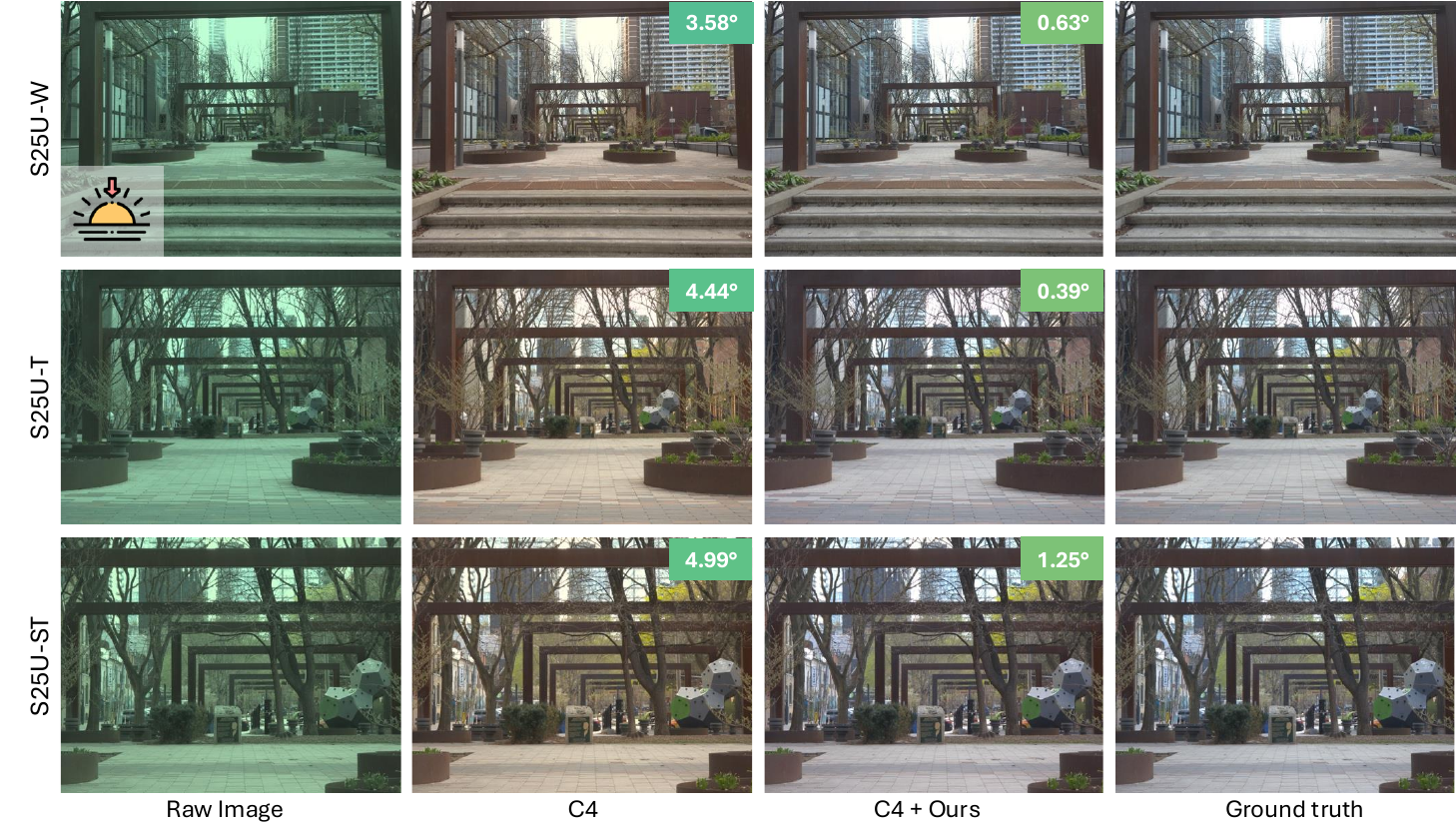}
\vspace{-5mm}
\caption{Qualitative results. Shown is a sunset scene captured by the S25U wide, telephoto, and super-telephoto sensors, white-balanced using C4 \cite{C4} predicted illuminant, C4 corrected with our mapping, and the ground truth aesthetically-preferred illuminant. \label{fig:qualitative-c4}}
\vspace{-2mm}
\end{figure*}

\begin{figure*}[t]
\centering
\includegraphics[width=\linewidth]{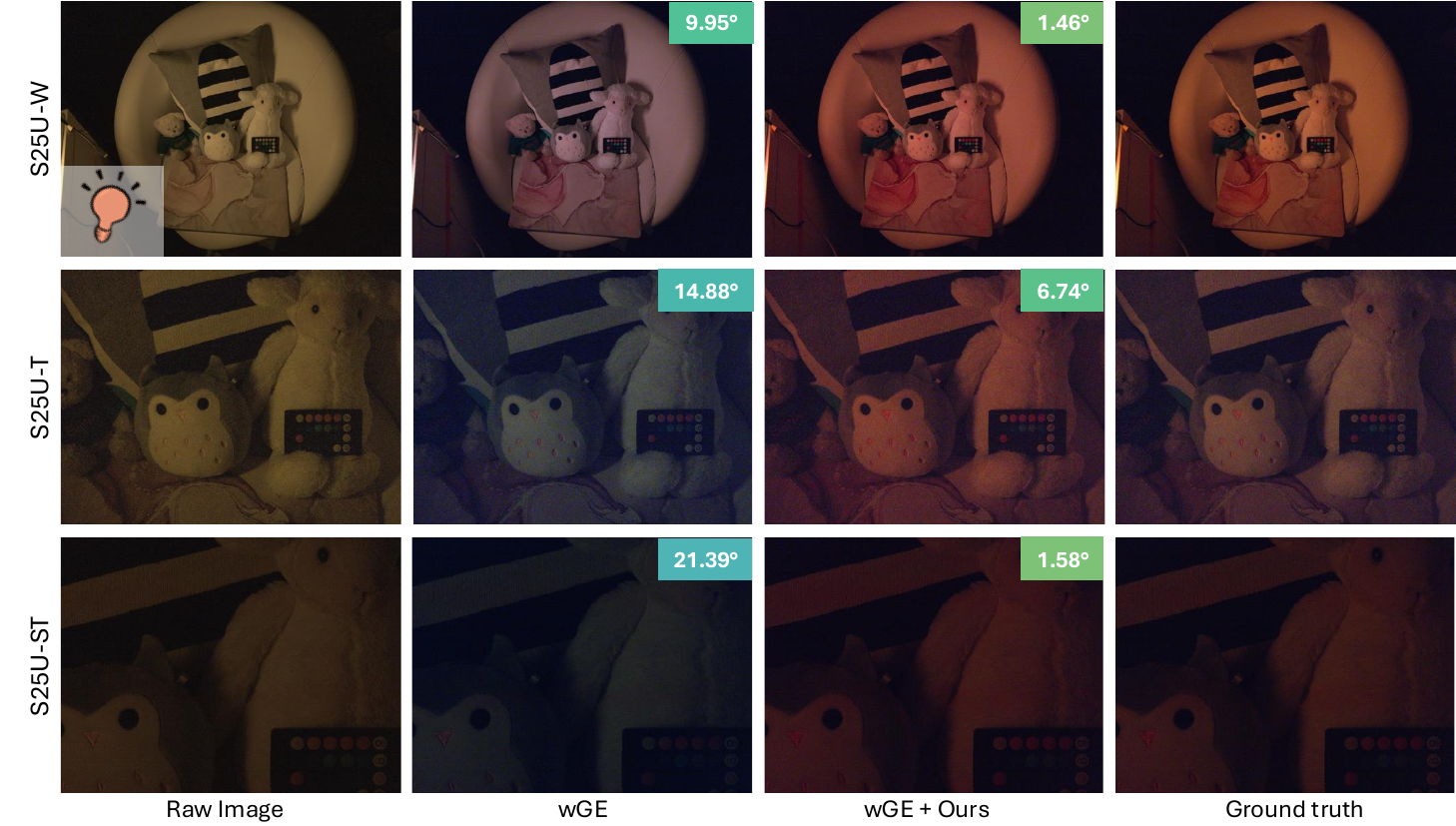}
\vspace{-5mm}
\caption{Qualitative results. Shown is an indoor scene illuminated by an orange LED light, captured by the S25U wide, telephoto, and super-telephoto sensors, white-balanced using wGE \cite{wGE} predicted illuminant, wGE corrected with our mapping, and the ground truth aesthetically-preferred illuminant. \label{fig:qualitative-wge}}
\vspace{-5mm}
\end{figure*}

\section{Qualitative Results}
\label{supp-sec:qualitative-results}
In the main paper, we presented qualitative results for our proposed mapping applied on cross-camera methods C5 \cite{C5} and gray world \cite{GW}. In this section, we present more results for our mapping based on C4 \cite{C4} in Fig.~\ref{fig:qualitative-c4}, and weighted gray-edge (wGE) \cite{wGE} in Fig.~\ref{fig:qualitative-wge}. As shown in the figures, our results match more closely with the ground-truth's aesthetic style than the cross-camera method's estimated neutral illuminant.

\newpage
{
    \small
    \bibliographystyle{ieeenat_fullname}
    \bibliography{ref}
}


\end{document}